\documentclass[10pt,journal,cspaper,compsoc]{IEEEtran}
\usepackage{times}
\usepackage{epsfig}
\usepackage{graphicx}
\usepackage{algorithm}
\usepackage{algorithmic}
\usepackage{amsmath}
\usepackage{amssymb}
\usepackage{mathrsfs}
\usepackage{subfig}
\usepackage{float}
\usepackage{multirow}
\usepackage{array}
\usepackage{color}
\usepackage{url}
\usepackage{booktabs}

\ifCLASSOPTIONcompsoc
\else
\fi
\ifCLASSINFOpdf
\else
\fi
\UseRawInputEncoding

\begin{document}
%
\title{Sparse Regularized Correlation Filter for UAV Object Tracking with adaptive Contextual Learning and Keyfilter Selection}
%
%
%
%

\author{Zhangjian Ji,~\IEEEmembership{Member,~IEEE,}
        Kai Feng, Yuhua Qian,~\IEEEmembership{Member,~IEEE  and Jiye Liang, Senior Member, IEEE}
\IEEEcompsocitemizethanks{\IEEEcompsocthanksitem Zhangjian Ji and Kai Feng are with the Key Laboratory of Computational Intelligence
and Chinese Information Processing of Ministry of Education, School of Computer and Information Technology, Shanxi University, Taiyuan 030006, China E-mail: \{jizhangjian@sxu.edu.cn, fengkai@sxu.edu.cn\} \protect\\

}
\IEEEcompsocitemizethanks{\IEEEcompsocthanksitem Yuhua Qian is with the Institute of Big Data Science
and Industry, the Key Laboratory of Computational Intelligence and
Chinese Information Processing of Ministry of Education, Shanxi University,
Taiyuan 030006, Shanxi, China
E-mail: \{ jinchengqyh@126.com \}
}
\IEEEcompsocitemizethanks{\IEEEcompsocthanksitem Jiye Liang is with the Key Laboratory of Computational Intelligence and Chinese
Information Processing of Ministry of Education, Shanxi University,
Taiyuan 030006, Shanxi, China
E-mail: \{ ljy@sxu.edu.cn \}
}
\thanks{}}

%
%

\markboth{Journal of \LaTeX\ Class Files,~Vol.~6, No.~1, Mar.~2022}%
{Shell \MakeLowercase{\textit{et al.}}: Bare Demo of IEEEtran.cls for Computer Society Journals}
%


\IEEEcompsoctitleabstractindextext{%
\begin{abstract}
Recently, correlation filter has been widely applied in unmanned aerial vehicle (UAV) tracking due to its high frame rates, robustness and low calculation resources. However, it is fragile because of two inherent defects, i.e, boundary effect and filter corruption. Some methods by enlarging the search area can mitigate the boundary effect, yet introducing the undesired background distractors. Another approaches can alleviate the temporal degeneration of learned filters by introducing the temporal regularizer, which depends on the assumption that the filers between consecutive frames should be coherent. In fact, sometimes the filers at the ($t-1$)th frame is vulnerable to heavy occlusion from backgrounds, which causes that the assumption does not hold. To handle them, in this work, we propose a novel $\ell_{1}$ regularization correlation filter with adaptive contextual learning and keyfilter selection for UAV tracking. Firstly, we adaptively detect the positions of effective contextual distractors by the aid of the distribution of local maximum values on the response map of current frame which is generated by using the previous correlation filter model. Next, we eliminate inconsistent labels for the tracked target by removing one on each distractor and develop a new score scheme for each distractor. Then, we can select the keyfilter from the filters pool by finding the maximal similarity between the target at the current frame and the target template corresponding to each filter in the filters pool. Finally, quantitative and qualitative experiments on three authoritative UAV datasets show that the proposed method is superior to the state-of-the-art tracking methods based on correlation filter framework.
\end{abstract}

}

\maketitle

\IEEEdisplaynotcompsoctitleabstractindextext

%
\IEEEpeerreviewmaketitle

\section{Introduction}
\label{sec:intro}
\IEEEPARstart{V}{isual} object tracking is an important research topic in the computer vision field, which is pervasively applied in unmanned aerial vehicle (UAV) with the flexible, autonomous and extendable aerial platform, e.g., target following\cite{Alpher05}, aerial cinematography\cite{Alpher06} and aircraft tracking\cite{Alpher07}. Despite great progress over the past decades, it is still a challenging problem to design a robust tracker for UAV platform because UAV tracking needs to meet some additional challenges (e.g., strong vibration, aspect ratio change, out-of-view, large relative movement, harsh calculation resources and limited power capacity) in addition to confronting some complex factors (full/partial occlusion, illumination variation, pose changes, background clutter, in/out-of-plane, complex motion and object blur) in generic object tracking. Here, we mainly investigate how to design a robust UAV tracking algorithm which can deal with these problems mentioned above.

In literature, although deep features\cite{Alpher08} or deep architecture\cite{Alpher09} can significantly improve the tracking robustness and accuracy, the complex convolution operations and huge parameter scale have impeded their practical application in UAV tracking. In contrast, correlation filer (CF) framework\cite{Alpher04} is widely adopted in UAV tracking owing to its high efficiency, robustness and low calculation resources. The speed of CF-based framework is dramatically raised by the fact that the circulant matrix is used to generate a large number of negative samples and it can be implemented by the fast Fourier transform (FFT) in the frequency domain. However, the boundary effects produced by the FFT will severely slow the discriminative ability of the learned filer model. To alleviate the boundary effects, several methods\cite{Alpher03,Alpher10,Alpher11,Alpher12} expand the search area, but the enlargement maybe introduce more context noise, which would distract the detection phase especially in situations of similar objects around.

Lately, the context-aware (CA) framework\cite{Alpher13} is introduced into the CF model to alleviate the context distraction for visual tracking. Generally speaking, CA framework uniformly samples four/eight fixed background patches around the tracked object and regards them as negative samples to suppress the response of context distraction during the filter learning . However, in real UAV videos, the positions of context distractors  usually vary over time. Therefore, if sampling the context patches on the fixed positions of each frame, it can not reveal the real background interference sources, and the tracking performance may be also affected. In addition, some approaches (e.g., STRCF\cite{Alpher14} and LADCF\cite{Alpher02}) introduced temporal regularizer into filer learning in order to mitigate the temporal degeneration of learned filers. Nevertheless, the construction of temporal regularizer rests on the assumption that filters between consecutive frames should be coherent. In fact, sometimes the filers at the ($t-1$)th frame is vulnerable to heavy occlusion from backgrounds, which caused that the assumption does not hold. Li \emph{et al.}\cite{Alpher15} selected the filers on the periodic keyframes to construct the temporal regularizer, which is possible to bring distraction when the tracking on the keyframes is not reliable.

To handle these issues above, in this paper, we propose a novel $\ell_{1}$ regularization correlation filer with adaptive contextual learning and keyfiler selection for UAV tracking (as shown in Fig.\ref{fig1}). The main contributions of this work are summarized as follows:
\begin{figure*}[htp]
\includegraphics[width=17.5cm,height=9.0cm]{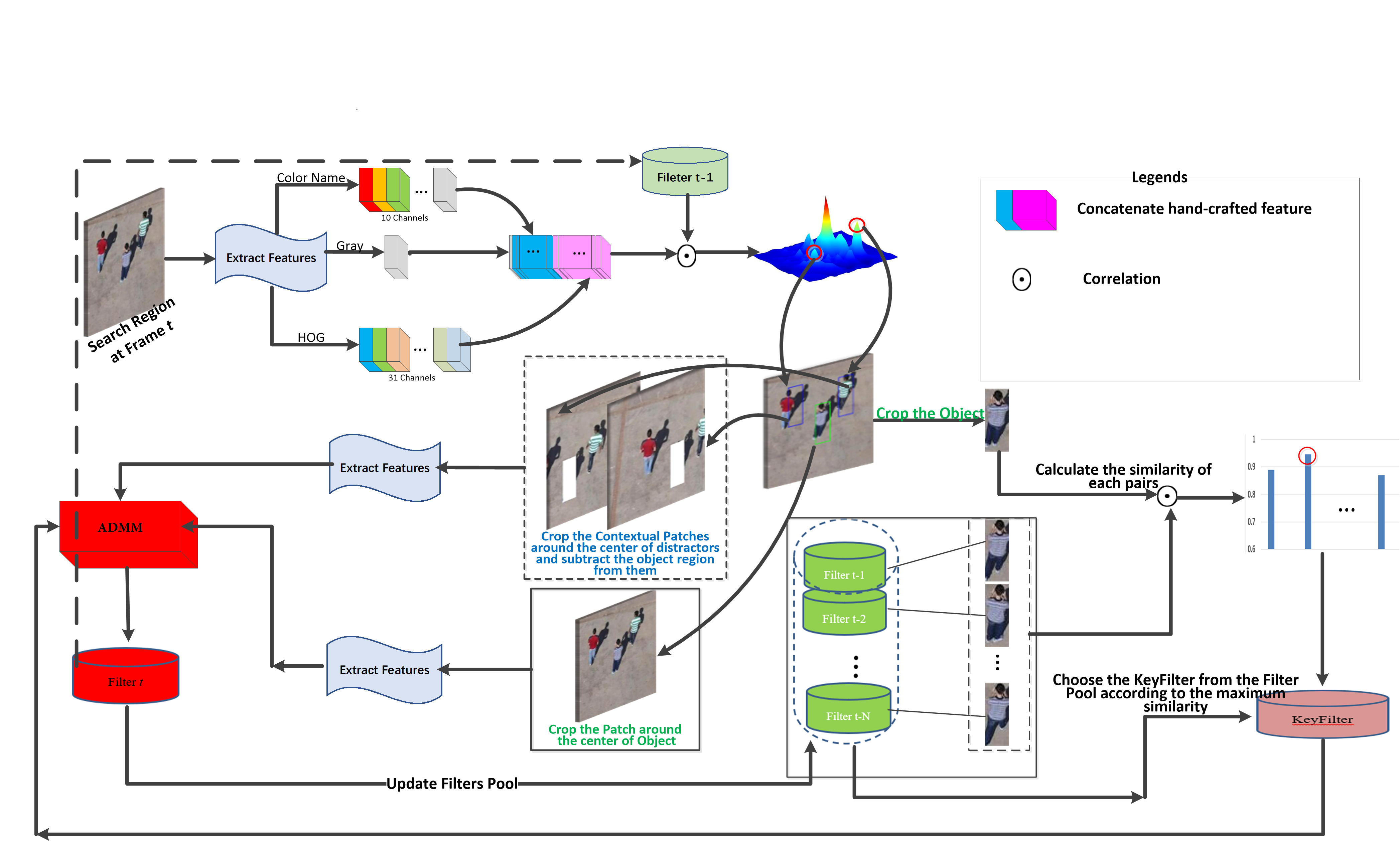}
   \caption{\small{The overall work-flow of the proposed tracking method. Given the searching region at $t$th frame, extract the color name (CN), gray and HoG features and calculate the response map with the learning filter at $(t-1)$th frame. Next, according to the distribution of maximum value and local maximums of response map, determine the positions of object and distractors, and crop the object patch and contextual patches from which the tracked object region is subtracted. Then, extract their features to feed into the solver of presented tracking model with the keyfilter that is chosen from the filters pool according to the maximum similarity between the object and object template corresponding to each filer in the filers pool. Finally, update the filters pool by the learning filter.}}
\label{fig1}
\end{figure*}
\begin{itemize}
  \item The true and effective distractors can be adaptively detected by the aid of the distribution of local maximum values on the response map of current frame which is generated by using the previous correlation filer model, and this scheme can alleviate the performance degradation of learned filers caused by the sampling false distractors at the fixed position of image frame.
  \item A new scoring scheme for each distractor is developed based on the product of its response peak and the normalized distance from its center to the tracked target's one.
  \item The label inconsistencies on the tracked target can be eliminated by removing it from each distractor.
  \item The keyfilter can be selected from the filters pool by finding the maximal similarity between the target at the current frame and the target template corresponding to each filter in the filters pool. Through the temporal restriction, CF degradation caused by occlusion or other issues can be mitigated.
  \item Quantitative and qualitative experiments on three authoritative UAV datasets show that the proposed method is superior to the state-of-the-art tracking methods based on correlation filter framework.
\end{itemize}

 The remaining of this work is organized as follows: Section \ref{sec:RW} introduces the related works. In Section \ref{sec:PF}, we elaborate on each component and optimization of the proposed method. Section \ref{sec:exp} shows the extensive evaluation and comparison of the proposed method and other state-of-the-art and the most related trackers on three well-known aerial tracking benchmarks. Finally, conclusions are explained in Section \ref{sec:con}

\section{Related works}\label{sec:RW}
\subsection{Tracking based on correlation filer}
Recently, CF-based framework is widely applied in visual tracking due to its hight computational efficiency\cite{Alpher04,Alpher08}. DCF\cite{Alpher04} is a representative work, which introduces multiple feature channels into their correlation filer tracking framework. Although DCF has good real time, accuracy and robustness, it can not effectively handle the various challenges, such as scale variation, spatial distortion, temporal degradation, etc.

To adapt the scale changes of the tracked object, DSST\cite{Alpher16} employs a separable 1D scale filter to estimate the scale of target. And Li \emph{et al.}\cite{Alpher17} adopt a multiple scale searching strategy to overcome the problem that traditional CF-based trackers can not deal with the scale variation. In order to mitigate the spatial distortion issue, SRDCF\cite{Alpher10} and its variant\cite{Alpher18,Alpher19} utilize a spatial regularization to penalize the filer coefficients outside the target bounding box during learning. However, due to their computational expense ($\thicksim 4$FPS) and the need for a lot of buffer memory, SRDCF and its variant are not a suitable choice for real time UAV tracking. The method of CFLB\cite{Alpher11} chooses a larger searching size and then crops the central patch of the signal that is same as the size of the filter by the binary matrix $\textbf{P}$ in each Alternating Direction Method of Multipliers (ADMM) iteration\cite{Alpher35}. But it was originally designed to learn from pixel intensities, which was shown to be inaccurate and worse generalization on challenging patterns\cite{Alpher20}. Therefore, to improve its performance, in\cite{Alpher12}, the CFLB is extended to multi-channel features to obtain a new approach (BACF). In\cite{Alpher03}, a multi-scale $\ell_{1}$ regularized correlation filer tracker (MSL1CFT) is proposed, which can learn a larger set of negative samples on larger image region for the training process, thereby obtaining a more discriminative model.

To alleviate the temporal degradation problem, some methods\cite{Alpher21,Alpher22,Alpher23} utilize good tracking results to update the model or reliable classifiers to redetect the target when the pre-defined conditions are satisfied. Under the assumption that the filers between the consecutive frames are coherent, STRCF\cite{Alpher14} and LADCF\cite{Alpher02} introduce the temporal regularization into learning correlation filer model for robust tracking. However, the above assumption exists a dilemma: the filers at the $(t-1)$th frame should adapt to significant deformation or resist to heavily occlusion. To solve this dilemma, in\cite{Alpher24}, Liang \emph{et al.} learn a transformation matrix from the initial filters to the previous filters so as to constrain the filters to be adaptive across temporal frames. Motivated by\cite{Alpher25}, Yan \emph{et al.}\cite{Alpher26} introduce a short-term template memory (STM) and a long-term template memory (LTM) into their correlation filter model to address the gradual model update.

\subsection{Tracking with contextual learning}
In the past decade, some works\cite{Alpher27,Alpher28,Alpher29,Alpher13} utilize the contextual information for better tracking performance. Yang \emph{et al.}\cite{Alpher28} use the auxiliary object in the context to promote long-time tracking. Nevertheless, similar objects in the context can still lead to tracking drift. In\cite{Alpher30}, Xiao \emph{et al.} exploit multi-level clustering to detect similar objects and other potential distractors in scenes. But all of these trackers do not generalize well, Mueller \emph{et al.}\cite{Alpher13} proposed a more generic model, in which the context patches around the target are explicitly integrated into the filter training as a penalty term, and it can be applied to most CF trackers\cite{Alpher17,Alpher31,Alpher32}. However, they sample context patches in the fixed position of background, which may not get the real background distractors in scenes. To solve this issue, in\cite{Alpher26}, background distractors are adaptively detected according to the response map of current frame which is generated by using the previous correlation filer model. But it does not consider that each distractor is different on the influence of filter learning.

\subsection{Tracking for Unmanned Aerial Vehicle}
Because of rapid movement, motion blur, limited computation capacity and intense vibration, UAV tracking is an extremely challenging task compared to the generic tracking. In literature, there are some satisfactory performance and real-time trackers on CPU. \emph{E.g}., Fu \emph{et al.}\cite{Alpher07} adopted a multi-resolution strategy to deal with the UAV large motion with a coarse-to-fine scheme. Huang \emph{et al. }\cite{Alpher33} proposed to restrict the response variation for repressing the aberrance happening in the detection phase. Zhang \emph{et al.}\cite{Alpher45}  adopt an environment-aware regularization term to learn the environment residual between two adjacent frames, which can enhance the discrimination and insensitivity of the filter in fickle tracking scenarios. A novel future-aware correlation filter tracker, which can efficiently exploit the contextual information of the upcoming frame to enhance its discriminative power, is proposed in \cite{Alpher44}. In addition, in contrast to the fixed spatial attention in \cite{Alpher10}, later works \cite{Alpher48,Alpher49}present a novel approach to online automatically and adaptively learn spatial attention mechanism by using target attention information \cite{Alpher49} or response variations \cite{Alpher48}. Different from traditional predefined label, in \cite{Alpher50}, Fu \emph{et al.} adopt an adaptive regression label to repress the distractors.
\section{Problem formulation and proposed framework}\label{sec:PF}
In this section, we present a novel $\ell_{1}$ regularization correlation filer tracker with adaptive contextual learning and keyfiler selection (L1CFT\_ACLKS). Since the proposed method considers the $\ell_{1}$ regularization correlation filer tracker (L1CFT)\cite{Alpher03} as our baseline tracker, we first briefly revisit it in Subsection \ref{sec:l1cft}. Then, in Subsection \ref{sec:ofpm}, the core object function of our proposed method is introduced in details. After that, we develop an efficient solution to optimize our core object function in Subsection \ref{sec:opt}, and employ it for object tracking in Subsection \ref{sec:ot}.
\subsection{Review of L1CFT}\label{sec:l1cft}
Given a training sample $\textbf{X}\in\mathbb{R}^{M\times N\times D}$ (i.e., the features with $M\times N$ pixels and $D$ channels extracted from the image patch $\textbf{I}\in \mathbb{R}^{lM\times lN}$, where $l$ is equal to the cell size used in the feature extraction.) and a Gaussian response map $\textbf{Y}\in\mathbb{R}^{M\times N}$. L1CFT \cite{Alpher03} trains a multi-channel discriminate correlation filter $\textbf{W}\in\mathbb{R}^{M\times N\times D}$ by minimizing the following cost function,
\begin{equation}\label{eq:1}
  \mathcal{E}(\textbf{W})=\frac{1}{2}||\sum_{d=1}^{D}\textbf{X}^{d}\ast\textbf{W}^{d}-\textbf{Y}||_{F}^{2}+\sum_{d=1}^{D}\sum_{m,n}\omega(m,n)|\textbf{W}^{d}|,
\end{equation}
Here, $\ast$ denotes the circular convolution operator, and $(m,n)\in\Omega :=\{0,\cdots,M-1\}\times\{0,\cdots,N-1\}$, which defines the same as the literature\cite{Alpher03}.

However, unfortunately, the formulation (\ref{eq:1}) does not have the closed solution and can only be solved iteratively via alternating direction method of multipliers (ADMM).
\subsection{Object function of the proposed method}\label{sec:ofpm}
In this subsection, we will elaborate our $\ell_{1}$ regularization correlation filer model with adaptive contextual learning and keyfiler selection.
\subsubsection{Formulation of spatial part}
In literature\cite{Alpher13}, it has been shown effectively to select contexts from the fixed position as negative samples. However, when locating target, it may be disturbed by background with similar response to the target coming from any possible position. Such distracors usually exist in different position, and are not always fixed around the target. In addition, sometimes there is no interference around the target. If we force to select some contexts as negative samples, it will only increase the complexity of correlation filter model but do not improve its performance. In order to solve these problems, we adopt an adaptive sampling method to detect background distractors.
\begin{equation}\label{eq:2}
  \begin{split}
  \mathcal{E}(\textbf{W})=&\frac{1}{2}||\sum_{d=1}^{D}\textbf{X}^{d}\ast\textbf{W}^{d}-\textbf{Y}||_{F}^{2}+\sum_{d=1}^{D}\sum_{m,n}\omega(m,n)|\textbf{W}^{d}|\\
  &\frac{1}{2}\sum_{p=1}^{P}\varpi_{p}||\sum_{d=1}^{D}\textbf{C}_{p}^{d}\ast\textbf{W}^{d}||_{F}^{2},
  \end{split}
\end{equation}
where $\textbf{C}_{p}\in\mathbb{R}^{M\times N\times D}$ represents the feature matrices of the $p$th background distractor detected by our algorithm and $\varpi_{p}$ is the weight of the $p$th patch measuring its repression power, which are described as follows.

\noindent\textbf{Adaptive distractors detection}: As is well-known, the discriminant performance of correlation filter model is fragile because of background contexts which are similar to it around the tracked target have the relative high responses on the response map. Thus, the procedure of distractor detection is as follows. First, based on the learned filter from the previous frame, we generate a response map on the search window. Second, we select several point with highest responses in the response map as the position of distractors. Finally, we set the sizes of distractors as the same with the target window and integrate them into correlation filter learning in our model.

\begin{figure}[htp]
\includegraphics[width=8.6cm,height=7.0cm]{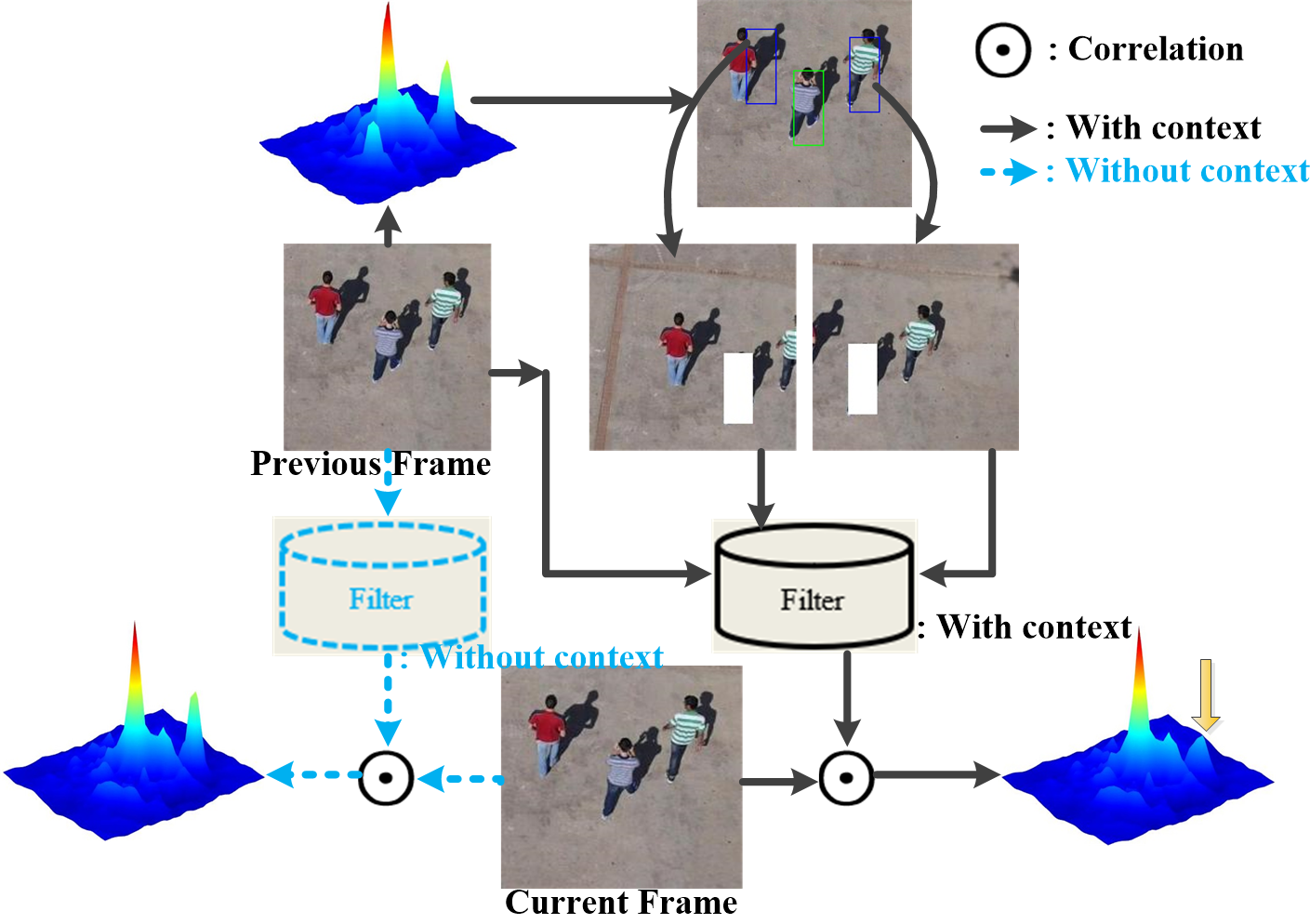}
   \caption{\small{Illustration of advantages of our adaptive contextual learning. The results show that our method can detect reasonable background distractors and effectively suppress their effects.}}
\label{fig:CA}
\end{figure}

Although all distractors have negative impact on target tracking, in order to reduce the extra redundancy in learning and heavy burden in circulation\cite{Alpher34}, we would not select distractors whose centers lie in range of bounding box for target and other distractors or whose responses on the response map are below a certain threshold, and only choose at most four representative distractors among candidates. In the detection phase, the actual influence of each distractor depends intuitively on its distance to the tracked object and its response value on the response map, so that we assign an adaptive weight to each distractor based on the product of its response peak and the normalized distance from its center to one of the tracker target during filter training (see the algorithm \ref{Alg:1} for details). Also, known for the formula (\ref{eq:2}), some non-zero Gaussian regression labels are assigned to the target region while the regression labels of each distractor are zeros, which would result in assigning  the inconsistent labels for the tracked target of searching region and the one that exists in each distractor patch. Therefore, we remove the tracker target region from each contextual patch when training filter, which makes our method repress the second highest peak of the response map more effectively (see the figure \ref{fig:CA}). In addition, in the whole procedure of collecting distractors, subwindow of few distractors may be out of view. In this situation, we will fill empty elements in subwindows with values of closest pixels on view boundary. The detailed procedure of collecting distractors is given in algorithm\ref{Alg:1}.
\begin{algorithm}
\hspace*{0.02in}\textbf{Input:} Current frame image: ${\rm F}_{now}\in \mathbb{R}^{hd \times wd}$, where $hd$ and $wd$ respectively denote the height and width of image; Maximum number of distractors: $P_{max}$; the number of distractors: $P$; the bounding box of object at current frame: $B_{obj}$; the response map at current frame: ${\rm R}\in \mathbb{R}^{hd \times wd}$; the threshold determining the distractor: $\epsilon$.\\
\textbf{Output:} bounding boxes of collecting distractors: $\mathcal{B}=\{B_{p}\}_{p=1}^{P}$ and the collecting background distractors set: $\mathcal{C}=\{{\rm C}_{p}\}_{p=1}^{P}$, the weight of detection distractors: $\{\varpi_{p}\}_{p=1}^{P}$
\begin{algorithmic}[1]
\STATE Initialize $P\leftarrow0$, $\epsilon\leftarrow 0.1$, $\mathcal{B}\leftarrow \Phi$, $\mathcal{C}\leftarrow \Phi$\\
\STATE Normalize the response map: $\rm R=\frac{R}{\max(R)}$\\
\STATE Generate the mask matrix ${\rm\Pi}\in\mathbb{R}^{hd \times wd}$, where
\begin{equation}\label{}
  {\rm\Pi}(x,y)=\left\{
  \begin{aligned}
   &0 , if\ (x,y)\ in\ ranges\ of\ B_{obj}\\
   &1
  \end{aligned}
  \right. \nonumber
\end{equation}
\\
\STATE The image ${\rm F_{rm}}$ that removes the tracked object from the current from is obtained by ${\rm F}_{rm}={\rm F}_{now}\circ{\rm\Pi}$, where $\circ$ denotes the Hadamard product operator.\\
\WHILE{$size(\mathcal{B})< P_{max}$}
{
\STATE  $\textbf{point} = find\_localmax\_response\_point({\rm R})$\\
 \IF {${\rm R}_{\textbf{point}}<\epsilon$}
   \RETURN $P$, $\{{\rm C}_{p}\}_{p=1}^{P}$
 \ELSE
   \IF{$\textbf{point}$ in ranges of $B_{p}$ or $B_{obj}$}
     \STATE ${\rm R}_{\textbf{point}}\leftarrow 0$
   \ELSE
     \STATE $P\leftarrow P+1$
     \STATE $B_{p}=generate\_boundingbox($\textbf{point}$,obj\_size)$
     \STATE ${\rm C}_{p}=generate\_distractor({\rm F}_{rm},\textbf{point},$ \\ $search\_size)$
     \STATE Calculate the Euclidean distance $dist_{p}$ between the center point of object and
            that of the $p$-th distractor.
     \STATE $\varpi_{p}={\rm R}_{\textbf{point}}\cdot \exp(-\frac{dist_{p}}{16})$
     \STATE ${\rm R}_{\textbf{point}}\leftarrow 0$
   \ENDIF
 \ENDIF
}
\ENDWHILE
\end{algorithmic}
\caption{Procedure of adaptive distractors detection}
\label{Alg:1}
\end{algorithm}

\noindent\textbf{Distractors update}: For the first frame, we construct our distractors set $\mathcal{C}$ by using the algorithm \ref{Alg:1} above.

In the subsequent frames, because of the change of the image scene and target position, the distractors in background will change obviously and the distractors collected from the previous frame are out of date and can not be used for current frame. Therefore, we will discard all the distractors acquired in the previous frame and re-detect the distractors in the current frame according to the algorithm \ref{Alg:1}.
\subsubsection{Formulation of temporal regularizer}\label{sec:FTR}
In literatures\cite{Alpher02,Alpher14}, in order to alleviate the degeneration of learned filters, the authors introduced the filter $\textbf{W}_{t-1}$ of the previous frame to construct the temporal regularizer. However, filter learning at the current frame will be heavily affected when the filter $\textbf{W}_{t-1}$ is corrupted by backgrounds and occlusions or is not adapt to target appearance variance during tracking.
To address this issue, we construct a candidate key-filter template set $\mathcal{W}=\{\textbf{W}_{pre}\}_{pre=1}^{L}$ and its corresponding object template set $\mathcal{T}=\{\textbf{t}_{pre}\}_{pre=1}^{L}$ according to the learning filer and cropping object template from the previous frame.
Next, we select the optimal reference filter from the candidate reference filter net by the similarity between the target at current frame and the target template corresponding to each reference filter in the object template set. Here, inspirited by\cite{Alpher25}, we also employ the convolution operation to measure the similarity between the templates. For convenience, we use $\textbf{f}_{i}$ to stand for corresponding feature vector of the template $\textbf{t}_{i}$. Therefore, if we want to measure the similarity between the template $\textbf{t}_{i}$ and $\textbf{t}_{j}$, we can calculate $\textbf{f}_{i}\ast \textbf{f}_{j}$. The more similar $\textbf{t}_{i}$ and $\textbf{t}_{j}$ are, the larger the value of $\textbf{f}_{i}\ast \textbf{f}_{j}$ is, and vice versa\cite{Alpher25}.

Based on above principles, we can calculate a $L$-dimensional similarity vector $\textbf{v}$ by the following formulate
\begin{equation}\label{eq:3}
  \textbf{v}=[\textbf{f}_{cur}\ast\textbf{f}_{1}, \textbf{f}_{cur}\ast\textbf{f}_{2}, \cdots,\textbf{f}_{cur}\ast\textbf{f}_{L}],
\end{equation}
where $\textbf{f}_{cur}$ denotes the feature vector of cropping the target in the current frame, $\textbf{f}_{1},\textbf{f}_{2}$ and $\textbf{f}_{L}$ respectively stand for corresponding to the feature vector of the object template $\textbf{t}_{1},\textbf{t}_{2}$ and $\textbf{t}_{L}$ in the target template set $\mathcal{T}$. Then, we can obtain the index $k$ of target template similar to the target of the current frame by
\begin{equation}\label{eq:4}
  k=\arg\max_{pre}(\textbf{v}).
\end{equation}

Thus, we can choose the filer $\textbf{W}_{k}$ of the candidate key-filters set as the keyfilter. After that, we can construct the temporal regularizer by our keyfilter $\textbf{W}_{k}$ and integerate it into our object function, i.e.,
\begin{equation}\label{eq:5}
  \begin{split}
  \mathcal{E}(\textbf{W})=&\frac{1}{2}||\sum_{d=1}^{D}\textbf{X}^{d}\ast\textbf{W}^{d}-\textbf{Y}||_{F}^{2}+\sum_{d=1}^{D}\sum_{m,n}\omega(m,n)|\textbf{W}^{d}|\\
  &\frac{1}{2}\sum_{p=1}^{P}\varpi_{p}||\sum_{d=1}^{D}\textbf{C}_{p}^{d}\ast\textbf{W}^{d}||_{F}^{2}+\frac{\lambda}{2}\sum_{d=1}^{D}||\textbf{W}^{d}-\textbf{W}_{k}^{d}||_{F}^{2},
  \end{split}
\end{equation}
where $\lambda$ is the controlling factor of the temporal regularizer term.

In order to adapt the target changes over time, we need to update the candidate key-filters set $\mathcal{W}$ and its corresponding object template set $\mathcal{T}$. If the sizes of $\mathcal{W}$ and $\mathcal{T}$ sets are lower than the predefined threshold, the learned filter and object template from current frame are collected into those two sets respectively. If not, we use them to replace the oldest filter and object template in those two sets. However, the filter $\textbf{W}_{1}$ and object template $\textbf{t}_{1}$ from the first frame will not take participate in updating. The main reason is that they are one of the most reliable representations of the target. To make the target recover from the occlusion or tracking failure at a certain extent, we will hold them in the whole process of tracking.
\subsection{Optimization}\label{sec:opt}
The formula (\ref{eq:5}) is a convex problem, which can be efficiently solved by ADMM\cite{Alpher35}. However, unfortunately, when we attempt to solve it by FFT in the frequency domain, a problem arises. The filter $\textbf{W}$ must be solved in the spatial domain because existing the $\ell_{1}$ constraint on $\textbf{W}$. Therefore, in order to solve the formula (\ref{eq:5}), we introduce an auxiliary variable $\textbf{U}$. In this case, the formula (\ref{eq:5}) can be equivalently expressed as
\begin{equation}\label{eq:6}
  \begin{split}
  \mathcal{E}(\textbf{W})=&\frac{1}{2}||\sum_{d=1}^{D}\textbf{X}^{d}\ast\textbf{W}^{d}-\textbf{Y}||_{F}^{2}+\sum_{d=1}^{D}\sum_{m,n}\omega(m,n)|\textbf{U}^{d}|+\\
  &\frac{1}{2}\sum_{p=1}^{P}\varpi_{p}||\sum_{d=1}^{D}\textbf{C}_{p}^{d}\ast\textbf{W}^{d}||_{F}^{2}+\frac{\lambda}{2}\sum_{d=1}^{D}||\textbf{W}^{d}-\textbf{W}_{k}^{d}||_{F}^{2}.\\
  & s.t. ~~~\textbf{W}^{d}=\textbf{U}^{d},~~~ \forall~ d=1,\cdots,D
  \end{split}
\end{equation}
After that, we employ the Augmented Lagrangian Method (ALM)\cite{Alpher35} to deal with the introduced equality constraints in Eq.(\ref{eq:6}) and rewrite it as:
\begin{equation}\label{eq:7}
  \begin{split}
  \mathcal{E}(\textbf{W})=&\frac{1}{2}||\sum_{d=1}^{D}\textbf{X}^{d}\ast\textbf{W}^{d}-\textbf{Y}||_{F}^{2}+\sum_{d=1}^{D}\sum_{m,n}\omega(m,n)|\textbf{U}^{d}|+\\
  &\frac{1}{2}\sum_{p=1}^{P}\varpi_{p}||\sum_{d=1}^{D}\textbf{C}_{p}^{d}\ast\textbf{W}^{d}||_{F}^{2}+\frac{\lambda}{2}\sum_{d=1}^{D}||\textbf{W}^{d}-\textbf{W}_{k}^{d}||_{F}^{2}\\
  & +\frac{\mu}{2}\sum_{d=1}^{D}||\textbf{W}^{d}-\textbf{U}^{d}+\textbf{V}^{d}||_{F}^{2},
  \end{split}
\end{equation}
where $\textbf{V}^{d}$ is the Lagrange multiplier and $\mu$ is the penalty factor that controls the rate of convergence of the ALM. Then, we may adopt the extension (ADMM) of ALM to alternatively optimize the following variables.

Optimizing $\textbf{W}^{d(i)}$. Because the $D$ channels of Eq.(\ref{eq:7}) are independent, we can decompose it into the $D$ independent subproblems. Given $\textbf{U}^{d(i-1)}$ and $\textbf{V}^{d(i-1)}$, in order to efficiently optimize $\textbf{W}^{d(i-1)}$, we employ the Parseval¡¯s theorem to solve the subproblem in the frequency domain:
\begin{equation}\label{eq:8}
  \begin{split}
  \hat{\textbf{W}}^{d(i)}=&\arg\min_{\hat{\textbf{W}}^{d}}\frac{1}{2}||\hat{\textbf{X}}^{d}\circ\hat{\textbf{W}}^{d}-\hat{\textbf{Y}}||_{F}^{2}+\\
  &\frac{1}{2}\sum_{p=1}^{P}\varpi_{p}||\hat{\textbf{C}}_{p}^{d}\circ\hat{\textbf{W}}^{d}||_{F}^{2}+\frac{\lambda}{2}||\hat{\textbf{W}}^{d}-\hat{\textbf{W}}_{k}^{d}||_{F}^{2}\\
  & +\frac{\mu}{2}||\hat{\textbf{W}}^{d}-\hat{\textbf{U}}^{d(i-1)}+\hat{\textbf{V}}^{d(i-1)}||_{F}^{2},
  \end{split}
\end{equation}
where $\hat{(\cdot)}$ represents the Discrete Fourier Transform (DFT), $\circ$ denotes the Hadamard product operator.

Let the derivative of Eq.(\ref{eq:8}) be zero, we can get the closed-form solution of $\hat{\textbf{W}}^{d(i)}$, i.e.,
\begin{equation}\label{eq:9}
\hat{\textbf{W}}^{d(i)}=\frac{(\hat{\textbf{X}}^{d})^{T}\circ\hat{\textbf{Y}}+\lambda\hat{\textbf{W}}_{k}^{d}+\mu(\hat{\textbf{U}}^{d(i-1)}-\hat{\textbf{V}}^{d(i-1)})}{(\hat{\textbf{X}}^{d})^{T}\circ\hat{\textbf{X}}^{d}+\sum_{p=1}^{P}\varpi_{p}(\hat{\textbf{C}}_{p}^{d})^{T}\circ\hat{\textbf{C}}_{p}^{d}+\lambda+\mu},
\end{equation}
where $\frac{\bullet}{\bullet}$ denotes the element-wise division.

In order to improve the calculational efficiency, we use the Sherman-Morrison formula\cite{Alpher37} to rewrite the Eq.(\ref{eq:9}). However, to utilize the Sherman-Morrison formula, $\sum_{p=1}^{P}\varpi_{p}(\hat{\textbf{C}}_{p}^{d})^{T}\circ\hat{\textbf{C}}_{p}^{d}$ should be broken into the product of two vectors $\sum_{p=1}^{P}\sqrt{\varpi_{p}}(\hat{\textbf{C}}_{p}^{d})^{T}$ and $\sum_{p=1}^{P}\sqrt{\varpi_{p}}(\hat{\textbf{C}}_{p}^{d})$ based on the assumption that it can be ignored when considering the difference
of omnidirectional contextual patches\cite{Alpher15}. Consequently, we can obtain the following solution:
\begin{equation}\label{eq:10}
\hat{\textbf{W}}^{d(i)}=\frac{1}{\lambda+\mu}(\textbf{I}+\frac{\hat{\textbf{Q}}\circ\hat{\textbf{Q}}^{T}}{\lambda+\mu+\hat{\textbf{Q}}^{T}\circ\hat{\textbf{Q}}})\boldsymbol{\Psi},
\end{equation}
where $\hat{\textbf{Q}}=\hat{\textbf{X}}^{d}+\sum_{p=1}^{P}\sqrt{\varpi_{p}}(\hat{\textbf{C}}_{p}^{d})$ and $\boldsymbol{\Psi}=(\hat{\textbf{X}}^{d})^{T}\circ\hat{\textbf{Y}}+\lambda\hat{\textbf{W}}_{k}^{d}+\mu(\hat{\textbf{U}}^{d(i-1)}-\hat{\textbf{V}}^{d(i-1)})$.

Optimizing $\textbf{U}^{d(i)}$. Given $\textbf{W}^{d(i)}$ and $\textbf{V}^{d(i-1)}$, we optimize the following sub-problem:
\begin{equation}\label{eq:11}
  \begin{split}
  \textbf{U}^{d(i)}=&\arg\min_{\textbf{U}^{d}}\sum_{d=1}^{D}\sum_{m,n}\omega(m,n)|\textbf{U}^{d}|+\\
  & +\frac{\mu}{2}\sum_{d=1}^{D}||\textbf{W}^{d(i)}-\textbf{U}^{d}+\textbf{V}^{d(i-1)}||_{F}^{2},
  \end{split}
\end{equation}
We may adopt the the shrinkage operator to obtain the closed-form solution for each element separately as follows:
\begin{equation}\label{eq:12}
 \textbf{U}^{d(i)}(m,n)=\mathscr{S}(\textbf{W}^{d(i)}(m,n)+\textbf{V}^{d(i-1)}(m,n),\frac{\omega(m,n)}{\mu})
\end{equation}
where $\mathscr{S}$ is the shrinkage operator:$\mathscr{S}(x,\beta)=sign(x)(|x|-\beta,0)$.

Update Lagrange Multiplier $\textbf{V}^{d(i)}$. Given $\textbf{W}^{d(i)}$ and $\textbf{U}^{d(i)}$, we update the Lagrange multiplier $\textbf{V}^{d(i)}$ by the following formula
\begin{equation}\label{eq:13}
\textbf{V}^{d(i)}=\textbf{V}^{d(i-1)}+\mu(\textbf{W}^{d(i)}-\textbf{U}^{d(i)}).
\end{equation}

Choice of $\mu$. A simple and common scheme\cite{Alpher35} for selecting ¦Ì is the following formula
\begin{equation}\label{eq:14}
\mu^{(i)}=\min(\mu_{max},\rho\mu^{(i-1)}),
\end{equation}
where $\mu_{max}$ is the maximum value of $\mu$ and $\rho$ is the scale factor. In our work, we found that setting $\mu^{(0)}=0.01$, $\rho=1.1$ and $\mu_{max}=10$ can obtain the best performance.


\subsection{Object tracking}\label{sec:ot}
In this subsection, we use the learned L1CFT\_ACLKS for object tracking. When obtaining the filter model $\textbf{W}_{\textmd{model}}$ from the previous frame, we first crop the search region centered at the estimated position in the previous frame and extract its $D$ channel features, denoted as $\textbf{Z}\in\mathbb{R}^{M\times N\times D}$, 
then calculate the response map as:
\begin{equation}\label{eq:15}
\boldsymbol{\Phi}=\mathscr{F}^{-1}(\sum_{d=1}^{D}\hat{\textbf{Z}}^{d}\circ\hat{\textbf{W}}_{\textmd{model}}^{d}),
\end{equation}
where $\mathscr{F}^{-1}(\cdot)$ denotes the inverse of Discrete Fourier Transform (IDFT).

However, because we don't use the cell size with $1\times 1$ pixels when extracting the features, $\boldsymbol{\Phi}$ is the response map on the coarser grid, by which our method only gets a inaccuracy target position. In order to improve the precision of position estimation, the response map $\boldsymbol{\Phi}$ is interpolated to the subpixel-dense ones by the approach in\cite{Alpher10}. After that, the position of target can be  estimated by searching for the maximum value in subpixel-dense response map. Meanwhile, to adapt the scale changes of target, we train the scale filter\cite{Alpher47} to estimate its scale variations. In addition, during tracking, the object appearance usually changes slowly because of some challenging factors (e.g., illumination variance and pose change). We adopt the same updating as the traditional DCF method\cite{Alpher36}:
\begin{equation}\label{eq:16}
\textbf{W}_{\textmd{model}}=(1-\alpha)\textbf{W}_{\textmd{model}}+\alpha\textbf{W},
\end{equation}
where $\alpha$ is the updating rate.


\section{Experiments}\label{sec:exp}
In this section, we first describe the implementation details, including feature representation and parameters settings used in our proposed method. Next, we introduce the benchmark datesets and evaluation metrics used in our proposed method in detail. Then, we conduct ablation study and parameter analysis. Finally, we compare our proposed method with some the most related state-of-the-art methods on different datasets and discuss the advantages of our method.
\subsection{Implementation details}
Our approach is implemented with native matlab2018b and all the experiments are run on a PC equipped with a 3.6GHZ Intel i7-4790CPU,8G RAM and a single NVIDIA GTX 1060 GPU.

\textbf{Feature representation:} Considering the real-time requirements for UAV tracking, we only use the hand-crafted features in our method and other trackers used in comparative experiments. Similar to other correlation filer-based tracking methods\cite{Alpher33,Alpher14,Alpher12}, for gray sequences, we adopt the 31-channel HOG features and gray features. However, for color sequences, the 31-channel HOG features, color name features with 10 dimentions and gray feature are utilized in our proposed method. In addition, to partly eliminate the boundary effect produced by the Fast Fourier Transform (FFT), each dimension features of the samples are multiplied by a cosine window.

\textbf{Parameters setup:} In our method, the HOG features adopt a cell size of $4\times 4$ pixels and each sample is represented by a
square $M\times N$ grid of cells (i.e. $M = N=50$). In order to ensure a maximum sample size of $M\times N$ cells, we crop the image region area of the samples with $4^{2}$ times the target area from the training image and resize it to $4M \times 4N$ pixels. we set the regularization parameter $\lambda=0.01$ in Eq.(\ref{eq:8}). The update rate $\alpha$ is set as $0.019$. All other parameters in this work have been given in the Sec.\ref{sec:PF}. For comparative analysis, we use the same parameter values and initialization for all the sequences. Other parameter
settings that not mentioned in the article are available in our source code\footnote{Our source code will be soon downloaded in our personal homepage:\url{https://sites.google.com/site/jzhang8455}} to be released for accessible reproducible research.

\subsection{Experimental setup}
\textbf{Datasets:} To assess the performance of our proposed method (L1CFT\_ACLKS), we perform the qualitative and quantitative experimental evaluations on the several public benchmark datasets: DTB70\footnote{\label{fn:2}The sequences together with the ground-truth and evaluation toolkit are available at: \url{https://github.com/flyers/drone-tracking}}\cite{Alpher38}, UAV123@10fps\footnote{\label{fn:2}The sequences together with the ground-truth are available at: \url{https://cemse.kaust.edu.sa/ivul/uav123}}\cite{Alpher39} and UAV112\cite{Alpher43}
, which include 305 challenging image sequences with together over 86000 frames.

\textbf{Evaluation metrics:} In order to an authoritative and objective evaluation, we follow the One Pass Evaluation (OPE) protocol used in \cite{Alpher41}. Thus, the center local error (CLE) and success rate (SR) based on the OPE are used to validate the performance of different methods quantitatively. Concretely, the CLE is computed as the average Euclidean distance between the estimated center location of the target and the ground-truth and SR is obtained by calculating the intersection over union (IoU) of the groundtruth and estimated bounding boxes. To vividly illustrate the results, the distance precision (DP) and overlap precision (OP) are adopted as main criteria\cite{Alpher41}. Specifically, the DP is the relative number of frames in the sequence where the center location error is smaller than a certain threshold (20 pixels in this work). The OP is defined as the percentage of frames where the bounding box overlap exceeds a threshold of 0.5, which corresponds to the PASCAL evaluation criterion\cite{Alpher42}. Except for the DP and OP metrics, the precision and success plots\cite{Alpher41} have also been adopted to measure the overall tracking performance. For the precision and success plots, we respectively use the DP value of each tracker and the area under curve
(AUC) score of each success plot to rank all trackers. Additionally, the speed of each tracker is measured by Frames per Second (FPS).

\subsection{Ablation study and parameter analysis}
Here, on the UAV123@10fps datasets\cite{Alpher39}, the target's motion between whose successive frames has become larger compared to original UAV123@30fps because it is downsampled from 30-frames/s to 10-frames/s, we deeply analysis the impact of each component in our proposed method (L1CFT\_ACLKS), i.e., adaptive contextual learning (ACL) and adaptive keyfiler selection (AKL). In order to more comprehensively validate the performance of our method, we study the effectiveness of the fixed-period keyfiler selection and fixed position intermittent contextual learning (FKS and FICL are adopted in KAOT\cite{Alpher15}) used in our method. In addition, we also give the corresponding results when the AKL component of our method is replaced by the traditional temporal regularization (TR is adopted in STRCF\cite{Alpher14}).  For this purpose, we designed six variants of L1CFT\_ACLKS: ``Baseline", ``Baseline+ACL", ``Baseline+AKS" , ``Baseline+FICL+AKS" , ``Baseline+ACL+TR" and ``Baseline+ACL+FKS". ``Baseline" refers to the original L1CFT tracker\cite{Alpher03} (see Sec.\ref{sec:l1cft}) using the same features as in L1CFT\_ACLKS. ``Baseline+ACL" denotes the adaptive contextual learning (ACL) is added to ``Baseline". ``Baseline+AKS" represents the tracker, where the adaptive keyfiler selection (AKL) is imposed on the ``Baseline". ``Baseline+FICL+AKS" denotes the tracker, where both FICL and AKS modules are added to ``Baseline". For the ``Baseline+ACL+TR",  the ACL and TR are appended to the ``Baseline". ``Baseline+ACL+FKS" is other tracker, which adds ACL and FKS components in ``Baseline".
\begin{table}
 \caption{\small{Analyzing the influence of the adaptive contextual learning (ACL) and adaptive keyfilter selection (AKL) on the DP/AUC scores (\%) of L1CFT\_ACLKS (Ours) on UAV123@10fps dataset. And compare the effect of replacing one of them with traditional temporal regularization (TR), the fixed-period keyfilter selection (FKS) and fixed-position intermittent contextual learning (FICL) respectively. The best results obtained by our method are highlighted by the \textcolor[rgb]{1.00,0.00,0.00}{red} bold.}}
  \centering
  \resizebox{.99\columnwidth}{!}{
  \begin{tabular}{ccccccc}
    \toprule
     & ACL & AKS & FICL &FKS&TR&UAV123@10fps \\
    \midrule
    Baseline & & & & & &61.1/45.1\\
    Baseline+ACL & \checkmark & & & & &63.8/47.2  \\
    Baseline+AKS & &\checkmark & & & &63.5/47.0  \\
    Baseline+ACL+FKS &\checkmark && &\checkmark & & 63.7/47.2 \\
    Baseline+ACL+TR &\checkmark & & & &\checkmark & 63.9/47.3 \\
    Baseline+FICL+AKS & &\checkmark &\checkmark & & &59.1/43.9 \\
    L1CFT\_ACLKS(Ours) &\checkmark &\checkmark & & & &\textcolor[rgb]{1.00,0.00,0.00}{\textbf{66.4}}/\textcolor[rgb]{1.00,0.00,0.00}{\textbf{48.4}} \\
    \bottomrule
  \end{tabular}
  }
  \label{tab:example}
\end{table}
Table\ref{tab:example} shows that the adaptive contextual learning (ACL) and adaptive keyfilter selection (AKL) can both enhance tracking performance. Compared to the "Baseline", we observe that the ``Baseline+ACL" and ``Baseline+AKS" can respectively obtain 2.7\%/2.1\% and 2.4\%/1.9\% performance gains on UAV123@10fps. We also notice that our method (L1CFT\_ACLKS) separately outperforms ``Baseline+ACL" and ``Baseline+AKS" by 2.6\%/1.2\% and 2.9\%/1.4\% in terms of the DP/AUC scores on UAV123@10fps because of the usage of ACL and AKS. Moreover, we can find that the DP/AUC scores of our method on UAV123@10fps raise by 5.3\%/3.3\% than ones of ``Baseline". Furthermore, seeing ``Baseline+ACL+FKS" and ``Baseline+ACL+TR" in Table \ref{tab:example}, we find that their DP/AUC scores are lower by 2.7\%/1.2\% and 2.5\%/1.1\% than ones of our method, and there is no significant difference in performance between them and ``Baseline+ACL", which suggest that the AKS can improve tracking performance more than the FKS and TR. Then, By comparing ``Baseline+FICL+AKS" with our method (L1CFT\_ACLKS) and ``Baseline", we can conclude that the FICL has a negative impact on the tracking performance, e.g. when substituting the ACL one of our method by the FICL module, the DP and AUC scores of the compared version (``Baseline+FICL+AKS" ) of our method drop by more than 7 and 4 percent points, even lower than "Baseline". In summary, we can draw two conclusions: 1) the ACL and AKS modules can consistently help to improve the tracking performance; 2) compared to the AKS module, the ACL one plays a more important role in boosting the tracking performance.

Then, on UAV123@10fps dataset, we further conduct the ablation study to analyze the influence whether subtract the tracked object region in the detecting contextual patches. To evaluate it conveniently, we denote the corresponding module variant as ``ACL+" when does not remove the tracked object region of the detecting distractor patches. Table \ref{Tb:2} gives the results of two different modules: ``ACL" and ``ACL+" adopted in our method. We observe that the DP and AUC scores of our method decrease by 2.9\% and 1.8\% when it uses the ``ACL+" module instead of the ``ACL" one. Known from the proposed model Eq.(\ref{eq:5}), some non-zero Gaussian regression labels are assigned to the tracked target and the regression lables of detected distractors are zeros, which would cause the labels assigned to it to be inconsistent when we don't remove the tracked target region of the detected contextual patches. Such contradiction should be the main reason that degrades the performance of our proposed model when the ``ACL+" module replaces ``ACL" one.
\begin{table}[!h]
\caption{\small{Comparing the results of two different modules: ``ACL" and ``ACL+" adopted in our method based on the distance precision (DP) and area under curve (AUC) scores. The entries in \textcolor[rgb]{1.00,0.00,0.00}{red} denote the best results.}}
\begin{center}
\begin{tabular}{ccc}
  \toprule
   & ACL+ & ACL \\
  \hline
  AUC score (\%) & 46.6 &\textcolor[rgb]{1.00,0.00,0.00}{\textbf{48.4}} \\
  DP score (\%)  & 63.5  &\textcolor[rgb]{1.00,0.00,0.00}{\textbf{66.4}}\\
  \bottomrule
\end{tabular}
\end{center}
\label{Tb:2}
\end{table}

\begin{figure*}[t]
   \centering
   \subfloat[]
   {
    \label{Fig:Epara}
   \includegraphics[scale=0.5]{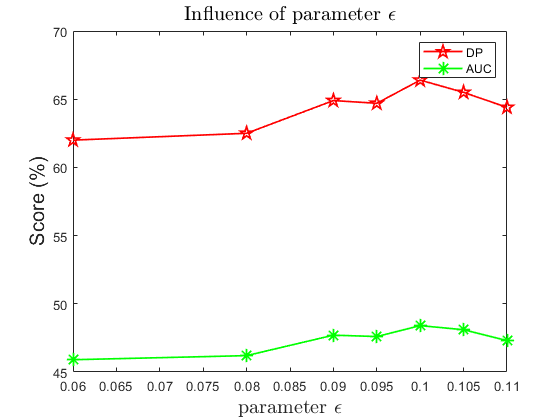}
   }
   \subfloat[]
   {
    \label{Fig:OP-OPE}
     \includegraphics[scale=0.5]{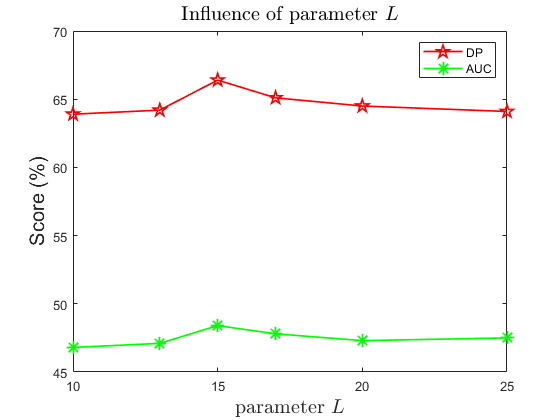}
   }
   \caption{\small{Influence of the $\epsilon$ and $L$ parameters on the DP and AUC scores on UAV123@10fps, respectively.}}
\label{Fig:para}
\end{figure*}

Finally, 
we also analyze the sensitivity of L1CFT\_ACLKS to the exclusive paramethers $\epsilon$ (the threshold to judge whether is the distractor) and $L$ (the size of candidate key-filter set and object template set). We set different values of $\epsilon$ and $L$ for L1CFT\_ACLKS and evaluate their influence on the DP and AUC scores of L1CFT\_ACLKS on UAV123@10fps. As shown in Figure \ref{Fig:para}(a), for the parameter $\epsilon$, the DP and AUC scores of L1CFT\_ACLKS fluctuate with respect to $\epsilon$ when it varies from 0.06 to 0.11 with a small step of 0.02. To obtain the value of parameter $\epsilon$ accurately, the step size is further reduced to 0.005 when approaching the estimated peak. When $\epsilon=0.1$, the DP and AUC scores of our method reach the highest point. Besides, as the value of parameter $\epsilon$ continues to increase, the performance becomes worse. Consequently, when the $\epsilon$ is set to 0.1, the L1CFT\_ACLKS can obtain the best tracking performance. This verifies that a proper threshold $\epsilon$, which is used to select the true distractors for our adaptive contextual learning scheme, plays a crucial role in the robust tracking. If $\epsilon$ is too large, fewer or no distractors will be chosen into our adaptive contextual learning scheme to learn the target tracking model, which would make the L1CFT\_ACLKS degenerate into the traditional correlation filer model and lose the ability to suppress the tracking drift caused by the distractors. If it's too small, more false distractors will be selected to learn the L1CFT\_ACLKS, which will greatly reduce the ability of the learned tracking model to distinguish the tracked target from the whole image scene. Figure \ref{Fig:para}(b) plots the curves of DP and AUC scores of L1CFT\_ACLKS when the parameter $L$ is set to 10 and increases in some certain steps \{3,2,2,3,5\} until 25. We observe that the L1CFT\_ACLKS achieves the best DP and AUC scores when $L=15$. This also validates that an appropriate size of candidate key-filters set is the extremely important for robust tracking. If $L$ is too small or too large, the performance of L1CFT\_ACLKS degrades significantly. The main reason is that the smaller $L$ lowers the diversity of candidate key-filter template set $\mathcal{W}$ which is not conductive to recovering from tracking failures, and the larger $L$ lets the candidate key-filter template set $\mathcal{W}$ and tracker object template set $\mathcal{T}$ buffer abundant redundancy and similar templates, which makes the key-filter that is chosen by the method given in subsection \ref{sec:FTR} potentially unreliable, which causes the tracking drift, besides adding the extra memory burden.
\subsection{Comprehensive evaluation on prevailing aerial tracking benchmarks}
Here, we compare our proposed tracking method with 12 state-of-the-art and the most related trackers from the literature: EMCF\cite{Alpher45}, FACF\cite{Alpher44}, AutoTrack\cite{Alpher48}, ARCF\cite{Alpher33}, LADCF\cite{Alpher02}, SRDCF\cite{Alpher10}, ECO\cite{Alpher08}, KCF\cite{Alpher04}, EFSCF, ASRCF\cite{Alpher46}, KAOT\cite{Alpher15} and DSST\cite{Alpher47} on UAV123@10fps, DTB70 and UAV112 datasets.
\subsubsection{Quantitative results of over performance}
The over evaluation between our method and these most related ones is reported in the Figure \ref{Fig:overall}, using the precise and success plots with DP and AUC scores in the figure legend. Seeing from the Figure \ref{Fig:overall}, our method obtains the most competitive performance on three aerial tracking benchmarks. More specifically, as shown in Figure \ref{Fig:overall}(a), on UAV123@10fps dataset, our method (L1CFT\_ACLKS) respectively obtains a gain of 0.4\% and 1.1\% in DP scores compared to the second and third trackers (EMCF\cite{Alpher45} and FACF\cite{Alpher46}), and in AUC score, our method also achieves competitive result, only 0.2\% lower than the best one (EMCF\cite{Alpher45}). Figure \ref{Fig:overall}(b) gives the over performance of ours method and 12 most related trackers on DTB70 dataset. We observe that our method has an advantage of 1.7\% over the second tracker (AutoTrack\cite{Alpher48}) on DP score, as well as a gain of 1.2\% over the second tracker (FACF\cite{Alpher44}) in terms of AUC. Again, compared with these most related trackers on UAV112 dataset, our L1CFT\_ACLKS significantly outperforms them in terms of DP and AUC with the scores of 69.7\% and 47.8\%, which is reported in the Figure \ref{Fig:overall}(c).

Besides, Table \ref{Tb:overall} demonstrates the overall evaluation of performance on all three datasets in terms of DP, AUC and speed. Seeing from it, we find that L1CFT\_ACLKS ranks No.1 among all the trackers, respectively with a gain of 1.2\% and 0.6\% than the second place FACF\cite{Alpher46} and EMCF\cite{Alpher45} in terms of DP and AUC scores. Notice that L1CFT\_ACLKS is superior to FACF by 0.7\% in AUC score and EMCF by 1.4\% in DP score, respectively. In short, L1CFT\_ACLKS shows a stable excellent performance in terms of DP and AUC, while others perform relatively unsteadily. Especially compared to the most similar KAOT\cite{Alpher15}, L1CFT\_ACLKS separately obtains a significant improvement of 3.5\% and 5.6\% in DP and AUC scores, as well as more than twice the average speed of KAOT, which should benefit from the usage of ``ACL" and ``AKS" modules because the ``ACL" can adaptively select the true distractors and the number of selected distractors is usually fewer, which can greatly reduce the computational complexity, compared to the ``FICL" used in KAOT, and the ``AKS" is able to choose the more reliable keyfilters than the ``FKS" adopted in KAOT, which can avoid introducing the distraction or filter degradation caused by the unreliable keyfilters (e.g., the learned filter is not reliable when the target on the keyframes is occluded or out of view).
\begin{figure*}[htp]
\centering
\includegraphics[scale=0.38]{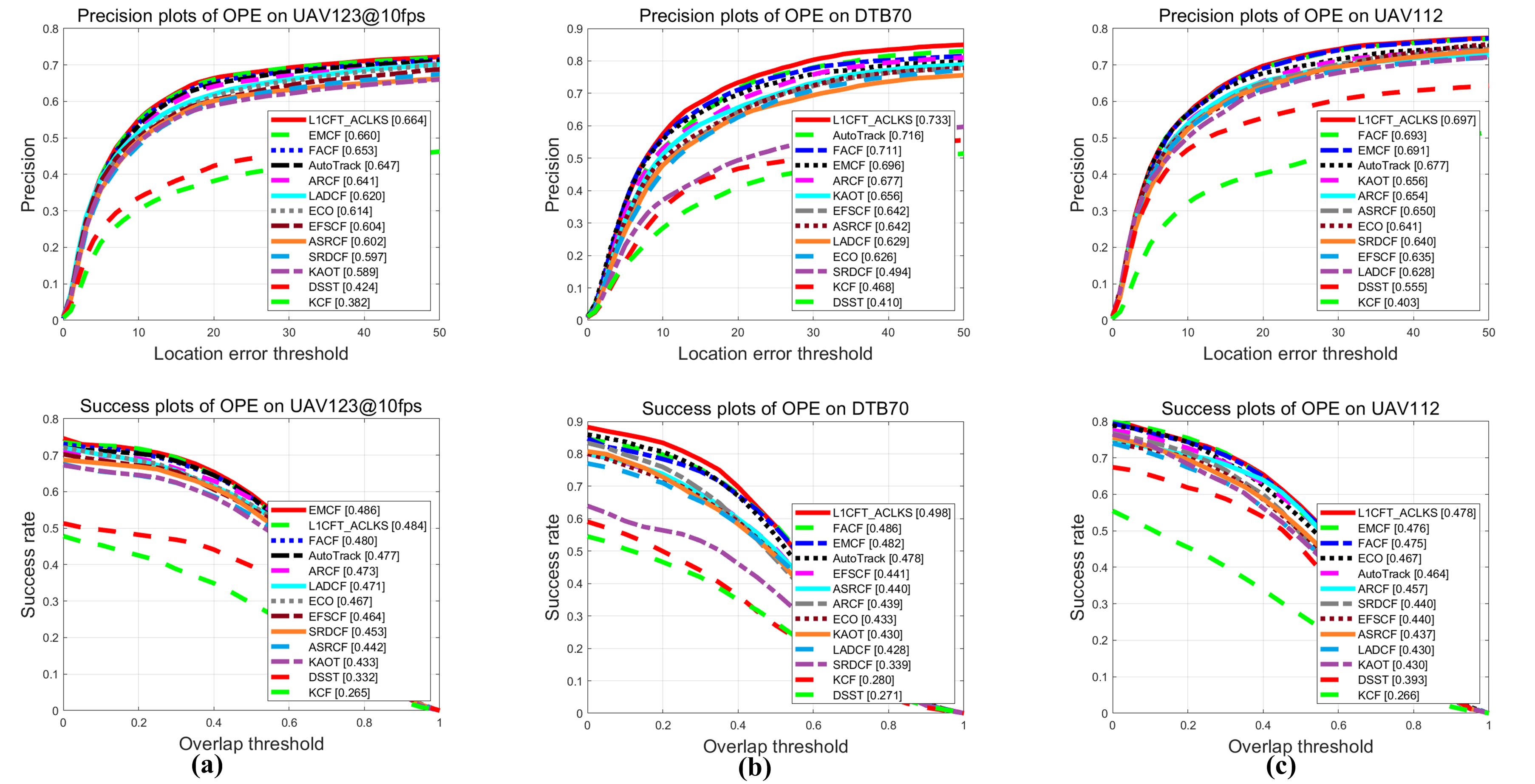}
\caption{Over performance comparison of our L1CFT\_ACLKS and 12 most related trackers on three well-known aerial tracking benchmarks: (a) UAV123@10fps\cite{Alpher39}, (b) DTB70\cite{Alpher38} and (c) UAV112\cite{Alpher43}. Our methods achieves competitive performance.}
\label{Fig:overall}
\end{figure*}
\begin{table*}[htp]
\caption{\small{Average performance of our L1CFT\_ACLKS and 12 most related trackers on three authoritative benchmarks: UAV123@10fps\cite{Alpher39}, DTB70\cite{Alpher38} and UAV112\cite{Alpher43}. The entries in \textcolor[rgb]{1.00,0.00,0.00}{red}, \textcolor[rgb]{0.00,1.00,0.00}{green} and \textcolor[rgb]{0.00,0.07,1.00}{blue} respectively denote the top three results.}}
\begin{center}
\resizebox{2\columnwidth}{!}{
\begin{tabular}{cccccccccccccc}
  \toprule
   & L1CFT\_ACLKS& EMCF&FACF &AutoTrack &ARCF & LADCF &ECO &EFSCF &ASRCF &SRDCF &KAOT & DSST &KCF\\
  \hline
  AUC score (\%) & \textcolor[rgb]{1.00,0.00,0.00}{\textbf{48.7}} &\textcolor[rgb]{0.00,1.00,0.00}{\textbf{48.1}}&\textcolor[rgb]{0.00,0.07,1.00}{\textbf{48.0}}&47.3&45.6&44.3&45.6&44.8&44.0&41.1&43.1&33.2&27.0 \\
  DP score (\%)  & \textcolor[rgb]{1.00,0.00,0.00}{\textbf{69.8}}  &\textcolor[rgb]{0.00,0.07,1.00}{\textbf{68.2}}&\textcolor[rgb]{0.00,1.00,0.00}{\textbf{68.6}}&68.0&65.7&62.6&62.7&62.7&63.1&57.7&63.4&46.3&41.8\\
  FPS & 28.7 &42.9&42.2&33.6&24.8&20.5&\textcolor[rgb]{0.00,0.00,1.00}{\textbf{48.5}}&17.5&21.9&5.8&13.5&\textcolor[rgb]{0.00,1.00,0.00}{\textbf{70.8}}&\textcolor[rgb]{1.00,0.00,0.00}{\textbf{401.7}}\\
  \bottomrule
\end{tabular}
}
\end{center}
\label{Tb:overall}
\end{table*}
\subsubsection{Quantitative results on various challenging sequence attributes}
\begin{table*}[htp]
\caption{\small{Average DP and AUC scores (\%) of our L1CFT\_ACLKS and 12 most related trackers under the common attributes of three authoritative datasets (UAV123@10fps\cite{Alpher39}, DTB70\cite{Alpher38} and UAV112\cite{Alpher43}), including ARC, CM, OV, SOB, SV and VC. The entries in \textcolor[rgb]{1.00,0.00,0.00}{red}, \textcolor[rgb]{0.00,1.00,0.00}{green} and \textcolor[rgb]{0.00,0.07,1.00}{blue} respectively denote the top three results.}}
\begin{center}
\resizebox{2.0\columnwidth}{!}{
\begin{tabular}{ccccccccccccc}
  \toprule
  \multirow{2}*{Tracker}&\multicolumn{2}{c}{ARC}&\multicolumn{2}{c}{CM}&\multicolumn{2}{c}{OV}&\multicolumn{2}{c}{SOB}&\multicolumn{2}{c}{SV}&\multicolumn{2}{c}{VC}\\
  \cline{2-13}
   &DP&AUC&DP&AUC&DP&AUC&DP&AUC&DP&AUC&DP&AUC\\
  \hline
  EMCF&57.7&\textcolor[rgb]{0.00,0.07,1.00}{\textbf{39.5}}&\textcolor[rgb]{0.00,0.07,1.00}{\textbf{63.5}}&\textcolor[rgb]{0.00,1.00,0.00}{\textbf{43.4}}&61.9&\textcolor[rgb]{0.00,1.00,0.00}{\textbf{42.4}}&\textcolor[rgb]{0.00,0.07,1.00}{\textbf{63.8}}&\textcolor[rgb]{0.00,1.00,0.00}{\textbf{43.4}}&\textcolor[rgb]{0.00,0.07,1.00}{\textbf{63.6}}&\textcolor[rgb]{0.00,1.00,0.00}{\textbf{43.7}}&63.7&\textcolor[rgb]{0.00,1.00,0.00}{\textbf{43.7}}\\
  FACF&\textcolor[rgb]{0.00,0.07,1.00}{\textbf{58.2}}&\textcolor[rgb]{0.00,0.07,1.00}{\textbf{39.5}}&63.2&\textcolor[rgb]{0.00,0.07,1.00}{\textbf{42.9}}&\textcolor[rgb]{0.00,0.07,1.00}{\textbf{62.0}}&\textcolor[rgb]{0.00,0.07,1.00}{\textbf{42.1}}&63.4&42.9&63.5&43.3&\textcolor[rgb]{0.00,0.07,1.00}{\textbf{63.9}}&\textcolor[rgb]{0.00,0.07,1.00}{\textbf{43.5}}\\
  AutoTrack&\textcolor[rgb]{0.00,1.00,0.00}{\textbf{59.2}}&\textcolor[rgb]{0.00,1.00,0.00}{\textbf{39.9}}&\textcolor[rgb]{0.00,1.00,0.00}{\textbf{63.8}}&\textcolor[rgb]{0.00,0.07,1.00}{\textbf{42.9}}&\textcolor[rgb]{0.00,1.00,0.00}{\textbf{63.0}}&\textcolor[rgb]{0.00,1.00,0.00}{\textbf{42.4}}&\textcolor[rgb]{0.00,1.00,0.00}{\textbf{64.3}}&\textcolor[rgb]{0.00,0.07,1.00}{\textbf{43.1}}&\textcolor[rgb]{0.00,1.00,0.00}{\textbf{64.4}}&\textcolor[rgb]{0.00,0.07,1.00}{\textbf{43.5}}&\textcolor[rgb]{0.00,1.00,0.00}{\textbf{64.3}}&\textcolor[rgb]{0.00,0.07,1.00}{\textbf{43.5}}\\
  ARCF&57.0&38.0&61.2&40.6&59.6&40.0&61.1&40.9&61.3&41.3&61.5&41.4\\
  LADCF&49.3&35.1&56.5&38.7&57.6&39.3&58.7&39.9&58.6&40.3&57.8&40.0\\
  ECO&52.1&37.5&57.6&40.6&58.0&40.7&59.0&41.3&58.7&41.5&58.3&41.3\\
  EFSCF&51.1&36.6&56.9&39.6&56.7&39.4&57.7&39.9&57.7&40.3&57.6&40.3\\
  ASRCF&52.1&35.8&57.6&39.2&57.1&38.8&58.2&39.4&58.4&39.9&58.4&39.9\\
  SRDCF&45.0&32.8&51.4&35.9&52.5&36.3&54.5&37.1&54.4&37.4&53.4&37.0\\
  KAOT&54.7&35.7&58.5&37.9&57.6&37.5&59.1&39.3&59.5&38.9&59.5&38.8\\
  DSST&36.1&27.4&39.5&28.2&40.4&28.5&42.1&29.3&42.5&29.9&41.6&29.5\\
  KCF&33.2&22.3&35.9&23.0&36.1&23.5&37.1&23.8&37.3&23.9&36.6&23.6\\
  \hline
  \textbf{L1CFT\_ACLKS}&\textcolor[rgb]{1.00,0.00,0.00}{\textbf{60.9}}&\textcolor[rgb]{1.00,0.00,0.00}{\textbf{40.9}}&\textcolor[rgb]{1.00,0.00,0.00}{\textbf{65.7}}&\textcolor[rgb]{1.00,0.00,0.00}{\textbf{44.0}}&\textcolor[rgb]{1.00,0.00,0.00}{\textbf{65.5}}&\textcolor[rgb]{1.00,0.00,0.00}{\textbf{43.9}}&\textcolor[rgb]{1.00,0.00,0.00}{\textbf{66.8}}&\textcolor[rgb]{1.00,0.00,0.00}{\textbf{44.6}}&\textcolor[rgb]{1.00,0.00,0.00}{\textbf{66.6}}&\textcolor[rgb]{1.00,0.00,0.00}{\textbf{44.9}}&\textcolor[rgb]{1.00,0.00,0.00}{\textbf{66.6}}&\textcolor[rgb]{1.00,0.00,0.00}{\textbf{44.8}}\\
  \bottomrule
\end{tabular}
}
\end{center}
\label{Tb:overall}
\end{table*}
\begin{figure*}[htp]
\centering
\includegraphics[scale=0.38]{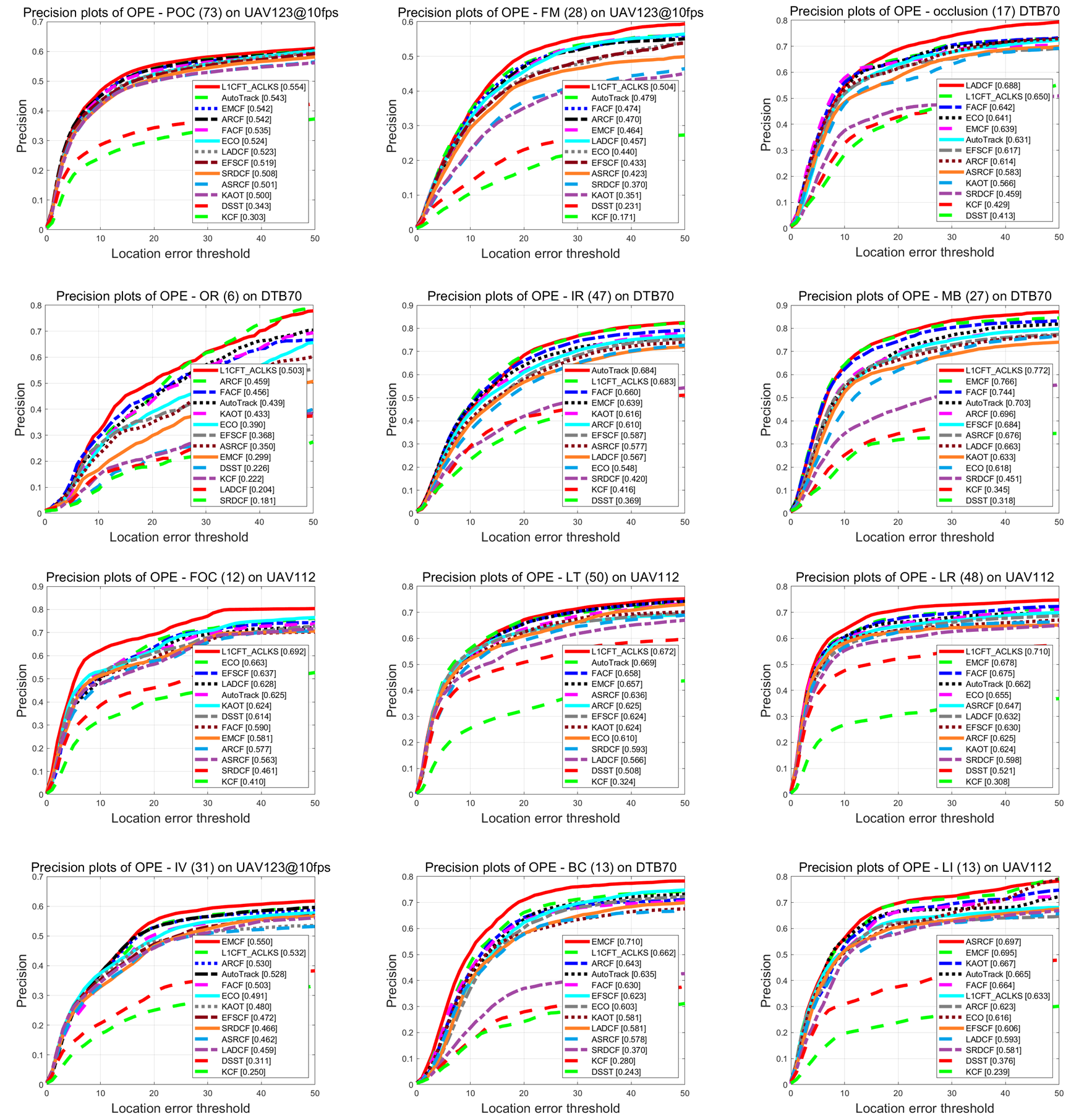}
\caption{Precision plots of our L1CFT\_ACLKS and 12 most related trackers on 12 different attributes (\emph{e.g.}, POC (Partial Occlusion), FM (Fast Motion), occlusion, OR (out-of-plane rotation), IR (in-plane rotation), MB (Motion Blur), FOC (Full Occlusion), LT (Long-term Tracking), LR (Low Resolution), IV (Illumination Variation), BC (Background Clutter) and LI (Low Illumination)) from three well-known aerial tracking benchmarks: (a) UAV123@10fps\cite{Alpher39}, (b) DTB70\cite{Alpher38} and (c) UAV112\cite{Alpher43}. The DP scores of each trackers are reported in the legends. Our methods achieves best performance on seven attributes and ranks second on four ones.}
\label{Fig:overall}
\end{figure*}
To rigorously analyze the performance of ours and other 12 most related trackers in different challenges, we first compare the proposed method (L1CFT\_ACLKS) and its competitors on six common attributes (\emph{e.g.}, ARC (Aspect Ratio Change), CM (Camera Motion), OV (Out of View), SOB (Similar Object), SV (Scale Variation) and VC (deformation (DTB70) and Viewpoint Change (UAV123@10fps and UAV112), because viewpoint affects target appearance significantly.) ) of three aerial tracking benchmarks, and the corresponding average DP/AUC scores of each tracker on different attributes are given in Table \ref{Tb:overall}. Seeing from it, we observe that our L1CFT\_ACLKS ranks first on average DP/AUC scores of all six common attributes. In case of similar object, our method outperforms the second tracker (AutoTrack) with an improvement of 2.5\% in terms of DP score and it surpasses the second place (EMCF) by 1.2\% in terms of AUC score, which is attributed to the fact that the ``ACL" module of L1CFT\_ACLKS can better suppress the response of distractors around the tracked target so as to learn a more discriminative model, which can distinguish the target from the distractors (similar object). In out-of view (OV) case, compared to the second AutoTrack (in terms of DP score) or EMCF (in terms of AUC score), our method obtains a gain of 2.5\% and 1.5\% respectively. The main reason is that the key-filer templates that our method buffers in $\mathcal{W}$ can help recover the correct track as quickly as possible when the target goes out of view and reappears.

In addition, Figure \ref{Fig:overall} describes the precision plots of our L1CFT\_ACLKS and 12 most related trackers on the remaining 12 attributes. Specifically, our method outperforms other trackers among the seven attributes (\emph{e.g.}, POC(0.554) and FM(0.504) on UAV123@10fps\cite{Alpher39}, OR(0.503) and MB(0.772) on DTB70\cite{Alpher38}, and FOC(0.692), LT(0.672) and LR(0.710) on UAV112\cite{Alpher43}) out of 12 for DP score. In the aberrant appearance change scenarios (Partial occlusion (POC) and Out-of-plane rotation (OR)), L1CFT\_ACLKS has a remarkable superiority of 1.1\% and 4.4\% compared to the second tracker. The main reason is that the adaptive selected keyfilter as temporal regularization can smoothly restrict the abnormal change of correlation filter and help it adapt to new appearance, thereby increasing the tracking robustness against various
appearance variation. For the occlusion sequences on DTB70\cite{Alpher38} and the full occlusion sequences on UAV112\cite{Alpher43}, our method obtains the second and first performance. Especially for the full occlusion situation,  L1CFT\_ACLKS is 2.9\% higher than the second tracker (ECO) in DP score because it can select the optimal keyfilter as a reference from  $\mathcal{W}$ by the similarity between the target of current frame and the target template corresponding to each keyfilter in target template set $\mathcal{T}$, which is helpful to recover the tracking timely when the target is no longer occluded fully. And more notably, compared with the most similar KAOT\cite{Alpher15}, its performance is significantly inferior to our L1CFT\_ACLKS (\emph{e.g.}, in the occlusion and full occlusion situations, our DP scores respectively exceeds its by 8.4\% and 6.8\%). The main reason is that the KAOT only adopts a simple periodic keyframe selection mechanism, which is impossible to introduce the distraction when the tracking on the keyframes is unreliable (\emph{e.g.}, the tracked target on the keyframe is occluded). In case of illumination variation (IV) and background clutter (BC), our approach is only second to the top tracker (EMCF) since its environment-aware model can better perceive temporal environment variation between consecutive frames than ours. Additionally, compared with other trackers, L1CFT\_ACLKS exhibits the best performance in the scenario of fast motion (FM), motion blur (MB), long-term tracking (LT) and low resolution (LR), which is desirable in aerial tracking. Unfortunately, for the low illumination (LI) attribute, the DP score of ours ranks sixth among ones of all trackers, 6.4\% lower than one of the top tracker (ASRCF). The reason may be that the low illumination affects the reliability of keyfilter selected by our method, thereby degenerating our tracking model.
\subsubsection{Qualitative results on challenging video sequences}
\begin{figure*}[htp]
\centering
\includegraphics[scale=0.63]{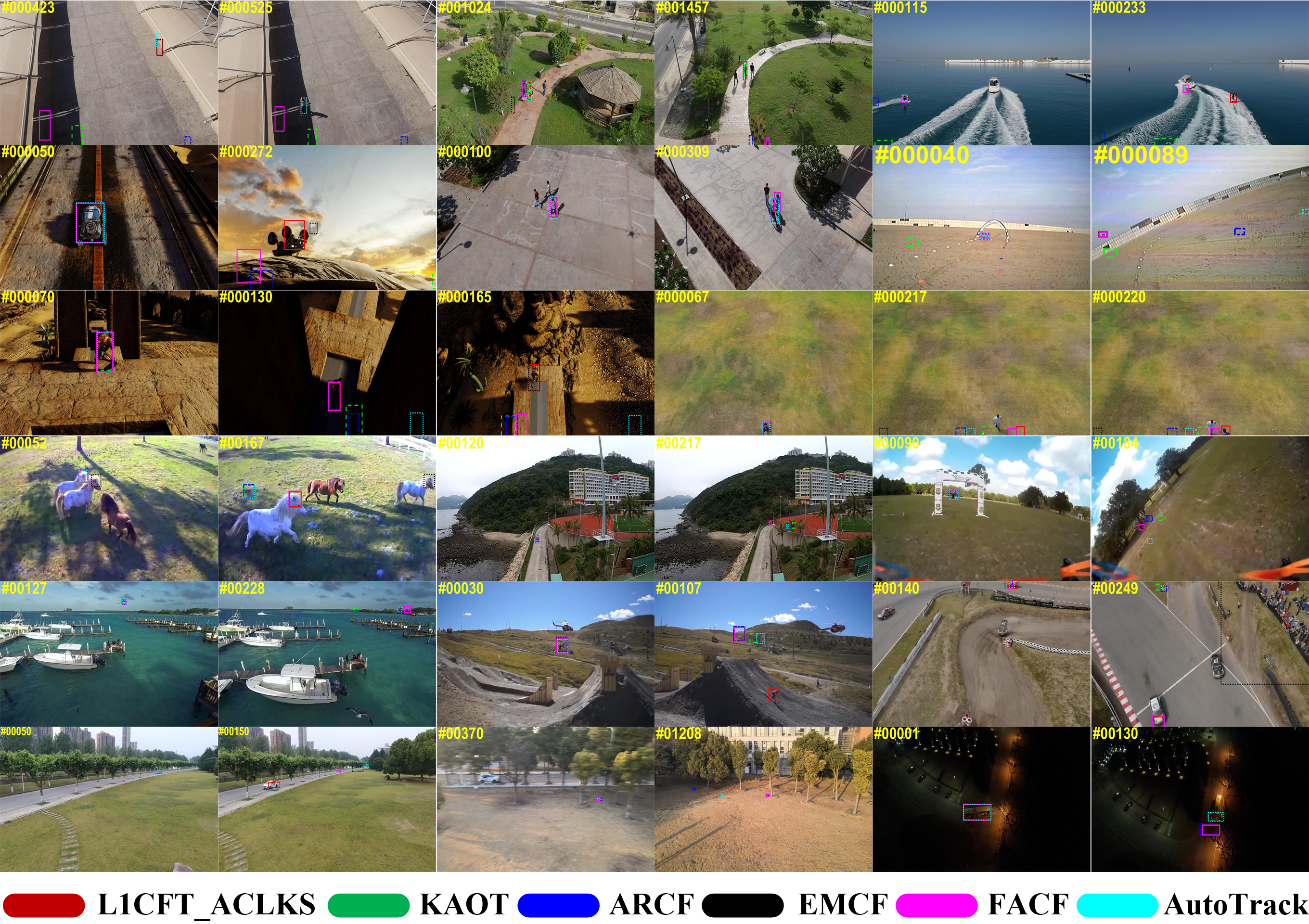}
\caption{Qualitative comparison of L1CFT\_ACLKS, KAOT and the rest of the top five trackers on some typically challenging sequences (from left to right and top to down are respectively person12\_2\_1, grpup3\_3\_1, wakeboard4\_1, car4\_s\_1, group1\_1\_1, UAV3\_1, person3\_s\_1 and person10\_1 from UAV123@10fps\cite{Alpher39}, Animal4\_1, car4\_1, chasingDrones1, Gull2\_1, MountainBike5\_1 and Racecar1\_1 from DTB70\cite{Alpher38} and car8\_1, UAV3\_2\_1 and bus2\_n\_1 from UAV112\cite{Alpher43}).
}
\label{Fig:qualitative}
\end{figure*}
The Figure \ref{Fig:qualitative} shows some qualitative results of our method, KAOT\cite{Alpher15} and the rest of the top five trackers on 17 challenging sequences. Ours performs well on most challenges, which benefits from the ``ACL" and ``AKS" modules that adopts in it. Ours can successfully tracked it without failures when the target is fully occluded (\emph{e.g.,} person12\_2\_1) or reappears after out of view (\emph{e.g.} person10\_1), which can be attributed to the fact that the key-filter templates buffering in $\mathcal{W}$ can help recover the correct tracking as quickly as possibly. When the similar object exists around the tracked target (\emph{e.g.}, grpup3\_3\_1, group1\_1\_1 and Animal4\_1), our method can obtain the better results than its competitors because our ``ACL" module can effectively repress the response of distractors around the tracked target, which can help learn a more discriminate model, especially good at handling with the similar object case. However, for low illumination sequence (\emph{e.g.}, bus2\_n\_1), our method fails to track the target at the 130th frame as it can not adapt the dark and low illumination.
\section{Conclusions}\label{sec:con}
In this work, we proposed a novel $\ell_{1}$ regularization correlation filter tracker with adaptive contextual learning and keyfilter selection to deal with challenge issues in UAV tracking. The adaptive contextual leaning can select the true distractors and the number of selected distractors is usually fewer, which alleviates the performance degradation of learned filters caused by the sampling patches at the fixed position of image frame that can not reveal the real interference, and greatly reduces the computational complexity (\emph{e.g.}, our method is more than twice as fast as the most related tracker (KAOT)). The adaptive keyfilter selection can avoid choosing the unreliable keyfilters (\emph{e.g.}, when the target on the keyframe is occluded or goes out of view.) to result in the model degradation. Results on three challenging aerial tracking benchmarks show that the overall performance of our tracker is superior to the state-of-the art tracking methods based on the correlation filter. Although our method performs best in most attributes, it still has some disadvantages, especially for low illumination. To handle this situation, one potential solution is to introduce the illumination adaption and anti-dark model into our tracker in future work.

%

%

\ifCLASSOPTIONcompsoc
  \section*{Acknowledgments}
\else
  \section*{Acknowledgment}
\fi

This work is supported by the National Natural Science Foundation of China Under Grant No. 61602288, Shanxi Provincial Natural Science Foundation of China Under Grant No. 20210302123443, Key Program of the National Natural Science Foundation of China (No. 62136005), National
Key Research and Development Program of China (No. 2020AAA0106100), and the 1331 Engineering Project of Shanxi Province. The authors also would like to thank the anonymous reviewers for their valuable suggestions.

\ifCLASSOPTIONcaptionsoff
  \newpage
\fi



%
{\small
\bibliographystyle{ieee}
\bibliography{egbib}
}
\begin{IEEEbiography}[{\includegraphics[width=1in,height=1.25in,clip,keepaspectratio]{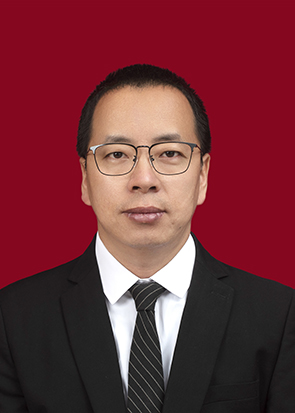}}]{Zhangjian ji}
received the B.S. degree from Wuhan University, China, in 2007, the M.E. degrees from the Institute of Geodesy and Geophysics, Chinese
Academy of Sciences (CAS), China, in 2010, and the Ph.D. degree in the University of Chinese Academy of Sciences, China, in 2015.

He is currently an Associate Professor at the School of Computer and Information Technology, Shanxi University, Taiyuan, China. His research interests include computer vision, pattern recognition, machine learning and human-computer interaction.
\end{IEEEbiography}
\begin{IEEEbiography}[{\includegraphics[width=1in,height=1.25in,clip,keepaspectratio]{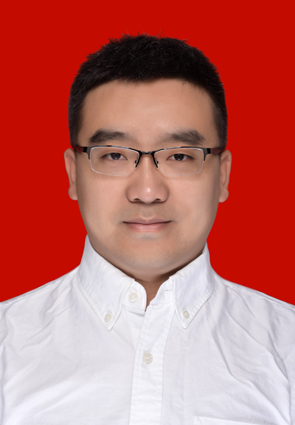}}]{Kai Feng}
received the PhD degree from Shanxi University, China, in 2014. He is currently an Associate Professor at the School of Computer and Information Technology, Shanxi University, China. His research interests include combinatorial optimization and interconnection network analysis.
\end{IEEEbiography}
\begin{IEEEbiography}[{\includegraphics[width=1in,height=1.25in,clip,keepaspectratio]{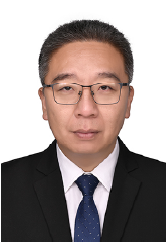}}]{Yuhua Qian}
is a professor and Ph.D. supervisor of Key Laboratory of Computational Intelligence and Chinese Information Processing of Ministry
of Education, China. He received the M.S. degree and the PhD degree in Computers with applications at Shanxi University in 2005 and 2011, respectively. He is best known for multigranulation rough sets in learning from categorical data and granular computing. He is actively pursuing research in pattern recognition, feature selection, rough set theory, granular computing and artificial intelligence. He has published more than 100 articles on these topics in international journals. He served on the Editorial Board of the International Journal of Knowledge-Based Organizations and Artificial Intelligence Research. He has served as the Program Chair or Special Issue Chair of the Conference on Rough Sets and Knowledge
Technology, the Joint Rough Set Symposium, and the International Conference on Intelligent Computing, etc., and also PC Members of many machine learning, data mining conferences.
\end{IEEEbiography}
\begin{IEEEbiography}[{\includegraphics[width=1in,height=1.25in,clip,keepaspectratio]{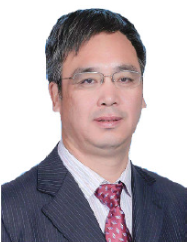}}]{Jiye Liang}
(Senior Member, IEEE) received the PhD degree from Xian Jiaotong University, Xian, China. He is currently a professor in Key Laboratory
of Computational Intelligence and Chinese Information Processing of Ministry of Education, the School of Computer and Information Technology,
Shanxi University, Taiyuan, China. His research interests include artificial intelligence, granular computing, data mining, and machine
learning. He has published more than 120 papers in his research fields, including the IEEE Transactions on Pattern Analysis and Machine Intelligence, IEEE Transactions on Knowledge and Data Engineering, IEEE Transactions on Fuzzy Systems, Data Mining and Knowledge Discovery and Artificial
Intelligence.
\end{IEEEbiography}
\end{document}